\documentclass{article} %
\usepackage{iclr2027_conference,times}

\usepackage{amsmath,amsfonts,bm}

\def\eqref#1{equation~\ref{#1}}
\def\Eqref#1{Equation~\ref{#1}}

\def\1{\bm{1}}

\def\rg{{\textnormal{g}}}

\def\ru{{\textnormal{u}}}

\def\rvepsilon{{\mathbf{\epsilon}}}

\def\rva{{\mathbf{a}}}

\def\rvg{{\mathbf{g}}}
\def\rvh{{\mathbf{h}}}

\def\rvx{{\mathbf{x}}}

\def\rvz{{\mathbf{z}}}
\newcommand{\valpha}{\boldsymbol{\alpha}}

\def\vzero{{\bm{0}}}
\def\vone{{\bm{1}}}
\def\vmu{{\bm{\mu}}}
\def\vtheta{{\bm{\theta}}}
\def\vpsi{{\bm{\psi}}}
\def\vphi{{\bm{\phi}}}
\def\vepsilon{{\bm{\epsilon}}}

\def\ve{{\bm{e}}}

\def\vh{{\bm{h}}}

\def\vq{{\bm{q}}}
\def\vr{{\bm{r}}}
\def\vs{{\bm{s}}}

\def\vu{{\bm{u}}}
\def\vv{{\bm{v}}}
\def\vw{{\bm{w}}}
\def\vx{{\bm{x}}}
\def\vy{{\bm{y}}}
\def\vz{{\bm{z}}}

\def\mA{{\bm{A}}}
\def\mB{{\bm{B}}}
\def\mC{{\bm{C}}}

\def\mI{{\bm{I}}}
\def\mJ{{\bm{J}}}

\def\mN{{\bm{N}}}

\def\mP{{\bm{P}}}

\def\mSigma{{\bm{\Sigma}}}

\DeclareMathAlphabet{\mathsfit}{\encodingdefault}{\sfdefault}{m}{sl}
\SetMathAlphabet{\mathsfit}{bold}{\encodingdefault}{\sfdefault}{bx}{n}

\newcommand{\E}{\mathbb{E}}

\newcommand{\R}{\mathbb{R}}

\newcommand{\Cov}{\mathrm{Cov}}

\usepackage{booktabs}   %
\usepackage{multirow}   %
\usepackage{hyperref}
\usepackage{url}
\usepackage{graphicx}
\usepackage{float}
\usepackage{caption}
\usepackage{pifont}
\usepackage{amsthm}
\newtheorem{theorem}{Theorem}
\newtheorem{lemma}{Lemma}
\newtheorem{proposition}{Proposition}
\newfloat{algorithm}{tbp}{loa}
\floatname{algorithm}{Algorithm}

\newcommand{\myparagraph}[1]{\vspace{0pt}\noindent{\bf #1}}

\title{\raggedright Self-supervised Collapse Prevention With Score-based Entropy Maximization}

\author{%
\textbf{Adarsh Kappiyath}$^{1}$ \quad
\textbf{Silpa Vadakkeeveetil Sreelatha}$^{1}$ \quad
\textbf{Xiatian Zhu}$^{1}$ \\
\textbf{Yann LeCun}$^{2,3}$ \quad
\textbf{Lu Yin}$^{4,1*}$ \quad
\textbf{Amir Bar}$^{5,3*}$ \\[0.4em]
\normalfont
$^{1}$University of Surrey \quad
$^{2}$New York University \quad
$^{3}$AMI Labs \\
$^{4}$Shenzhen University of Advanced Technology \quad
$^{5}$Imperial College London
}

\iclrpreprint
\begin{document}

\maketitle
{\renewcommand{\thefootnote}{*}\footnotetext{Equal advising.}}

\begin{abstract}

Diffusion models have demonstrated the effectiveness of score-based objectives for generative modeling, yet their potential for self-supervised representation learning remains underexplored. We introduce Entropy-Based Regularization (EBReg), a framework that uses latent score estimation to regularize the latent distribution during self-supervised training. EBReg jointly learns an encoder and a latent score model, and utilizes the learned score to provide a diversity-preserving signal that prevents representational collapse. Motivated by a mutual-information lower bound, we formulate the objective as view alignment with smoothed marginal entropy maximization and optimize the marginal entropy without explicit density evaluation. The same score model can additionally be used at test time to refine corrupted representations toward the training latent distribution. Empirically, EBReg learns competitive representations on ImageNet-1K, achieving 75.57\% linear-probe accuracy with ViT-L. Across diverse architectures, score-guided test-time adaptation further improves robustness on ImageNet-C, yielding an average relative accuracy gain of approximately 5\%.
\end{abstract}

\section{Introduction}
\label{sec:intro}

Diffusion models \citep{pmlr-v37-sohl-dickstein15,song2019generative,ho2020denoising} have established score estimation as a central component of modern generative modeling. However, representations learned through standard diffusion objectives often remain weaker than those produced by dedicated self-supervised learning (SSL) methods on recognition benchmarks \citep{yu2025repa,chen2024deconstructing}. Conversely, recent work shows that strong pretrained representations obtained from self-supervised learning can substantially improve diffusion training and generation, as demonstrated by REPA \citep{yu2025repa} and Representation Autoencoders \citep{zheng2026rae}. Earlier work on denoising autoencoders also demonstrated that denoising objectives can improve representation learning \citep{vincent2008extracting}. Together, these observations motivate the question: Can a score estimator and a self-supervised encoder be learned jointly in an end-to-end manner?

A central challenge in SSL is to enforce invariance across augmented views while maintaining sufficient diversity in the latent distribution to avoid collapse. Existing methods achieve this through different mechanisms, including negative-pair interactions \citep{chen2020simclr}, moment-based constraints \citep{bardes2022vicreg}, and regularization toward prescribed distributions \citep{balestriero2025lejepaprovablescalableselfsupervised}. Despite their different forms, these mechanisms preserve representation diversity while enforcing invariance. We investigate whether an analogous anti-collapse signal can be obtained from a latent score estimated via score matching.

\begin{figure}[t]
    \centering
    \includegraphics[width=0.85\linewidth]{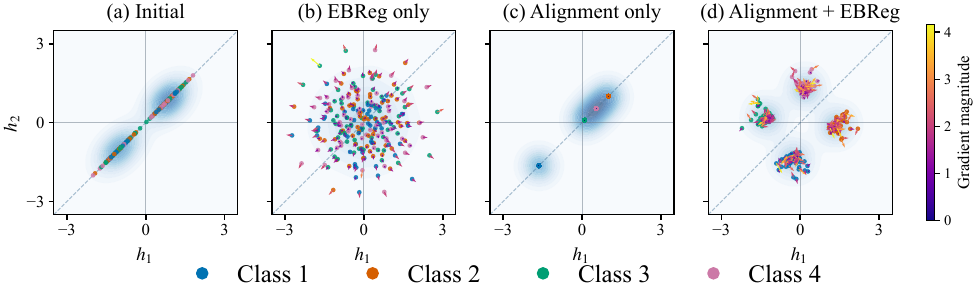}

    \includegraphics[width=0.66\linewidth]{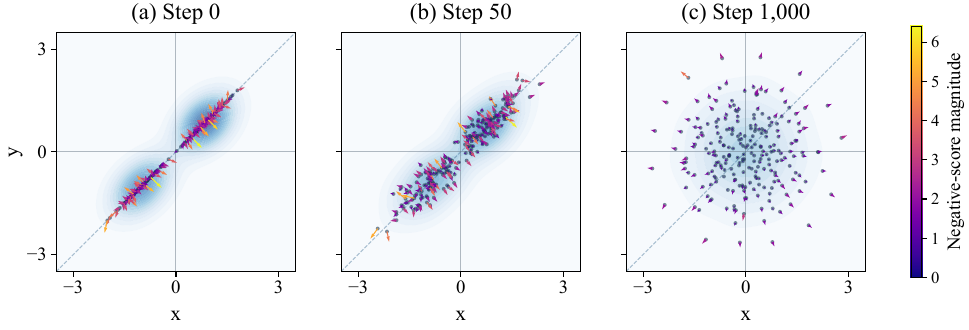}
   \caption{\textbf{Illustration of the EBReg anti-collapse mechanism.}
\textbf{Top:} (a) Representations lie near a one-dimensional subspace of $\mathbb{R}^2$. (b) EBReg alone spreads them across both dimensions. (c) Alignment alone contracts them, resulting in dimensional collapse. (d) Alignment + EBReg yields compact, separated classes while preventing dimensional collapse.
\textbf{Bottom:} Evolution of the latent distribution under EBReg. Samples are moved away from the diagonal and populate both dimensions, ultimately forming an isotropic Gaussian distribution.}
    \label{fig:negative-score-demo}
\end{figure}

We propose Entropy-Based Regularization (EBReg), a self-supervised framework that jointly trains an encoder and a latent score model. EBReg uses the score of the evolving latent distribution as an adaptive entropy-promoting signal. Figure~\ref{fig:negative-score-demo} illustrates this mechanism directly: samples concentrated along a low-dimensional structure are driven by the negative score toward a higher-entropy configuration that occupies both dimensions. Motivated by a variational mutual-information lower bound \citep{im}, the objective combines view alignment with maximizing the entropy of a Gaussian-smoothed latent distribution under a scale constraint. Since the latent density is unknown, EBReg estimates the score of the smoothed distribution via denoising score matching and uses it to optimize a gradient surrogate of marginal entropy without explicit density evaluation. The resulting encoder update promotes diversity in the latent distribution and counteracts representational collapse.

We further connect the EBReg formulation to existing anti-collapse regularizers \citep{balestriero2025lejepaprovablescalableselfsupervised, chen2020simclr} while replacing explicit negative pairs or fixed distribution matching with an adaptive score field. We characterize how latent scale constraints induce maximum-entropy solutions and relate these solutions to distributional targets and moment conditions underlying existing SSL regularizers. We further analyze the exact-score dynamics near dimensional collapse, including the behavior of low-variance directions. The learned score additionally enables test-time latent refinement, moving corrupted representations toward higher-density regions of the latent distribution while adapting only normalization parameters.

Empirically, EBReg learns competitive representations on ImageNet-1K \citep{imagenet} across convolutional and transformer backbones, performing comparably to established self-supervised methods including VICReg \citep{bardes2022vicreg} and LeJEPA \citep{balestriero2025lejepaprovablescalableselfsupervised} under the same pretraining and evaluation setup. Score-guided test-time refinement further improves robustness on ImageNet-C \citep{hendrycks2019benchmarking} across diverse architectures.

Our main contributions are summarized as follows:

\myparagraph{Method.} We derive an alignment--entropy objective from a variational mutual-information lower bound and introduce EBReg, which uses a learned latent score as an anti-collapse regularizer by optimizing representation entropy without explicit density evaluation or negative pairs.

\myparagraph{Theoretical analysis.} We analyze the exact-score dynamics near dimensional collapse and characterize when EBReg amplifies low-variance directions. We further show how latent constraints determine maximum-entropy solutions that recover distributional targets and moment conditions underlying existing SSL regularizers.

\myparagraph{Score-guided test-time adaptation.} We reuse the learned score as a label-free adaptation signal, updating only encoder normalization parameters to guide corrupted representations toward higher-density regions of the source latent distribution and improve robustness on ImageNet-C.

\section{Related Work}
\label{sec:rw}

\myparagraph{Latent Space Information Maximization.}
I-MAX \citep{Becker_1992}, an early precursor to joint-embedding methods, maximizes mutual information between module outputs for related inputs. Early tractable formulations use stochastic binary outputs or Gaussian assumptions for scalar continuous outputs. Predictability Maximization \citep{Schmidhuber_1993} instead combines agreement between paired representations with a discrimination loss that discourages collapse but does not in general yield a mutual-information objective. EBReg revisits the information-maximization principle through an alignment--entropy objective derived from an MI lower bound (Section~\ref{sec:objective}).  Unlike these earlier formulations, EBReg uses a learned score to optimize marginal entropy without explicit density or mutual-information estimation.

\myparagraph{Contrastive methods.}
\citet{NIPS1993_288cc0ff} introduced Siamese networks, which use shared-weight encoders to learn representations by distinguishing matching from non-matching pairs. Building on this principle, modern contrastive methods \citep{chen2020simclr,he2020moco,pmlr-v139-radford21a} make use of negative examples to provide a non-parametric repulsive signal that prevents representational collapse. Random negative sampling, however, can yield many uninformative
pairs in high-dimensional embedding spaces \citep{wu2017sampling}, resulting in dimensional collapse \citep{jing2022understanding}. Prototype-based methods such as SwAV, DINO, and MSN \citep{caron2020swav,caron2021dino,assran2022msn} avoid instance-level negatives by balancing assignments over a finite set of learned prototypes, which imposes an explicit discretization of the latent structure. EBReg instead learns a parametric score field to provide entropy-promoting directions without explicit negative comparisons. This design draws on the empirical success of score-based generative models \citep{song2021scorebased,rombach2022latentdiffusion} and theoretical evidence that score estimation can adapt to low-dimensional subspace structure, mitigating the curse of high dimensionality \citep{pmlr-v202-chen23o}.

\myparagraph{Non-contrastive methods.}
BYOL \citep{grill2020byol} helped revive non-contrastive learning through asymmetric branches and stop-gradient operations. Subsequent teacher--student distillation methods \citep{Assran_2023_CVPR,murlabadia2026vjepa2_1,oquab2024dinov} learn strong representations, with collapse avoidance governed by interactions among architecture, distillation, and optimization. A complementary approach explicitly regularizes representation statistics: VICReg \citep{bardes2022vicreg} enforces variance floors and penalizes cross-dimensional covariance, while LeJEPA \citep{balestriero2025lejepaprovablescalableselfsupervised} targets an isotropic Gaussian, with related distributional approaches in VISReg \citep{wu2026visreg} and Rectified LpJEPA \citep{kuang2026rectifiedlpjepajointembeddingpredictive}. EBReg instead maximizes smoothed marginal entropy subject to a scale constraint, allowing the entropy objective and constraint to determine the preferred distribution. EbReg under different scale constraints yields maximum-entropy targets that satisfy moment conditions used by existing methods (Table~\ref{tab:maxent-targets}).

\myparagraph{Reconstruction based methods.}
Another prominent representation-learning paradigm uses reconstruction objectives \citep{hinton2006reducing,vincent2008extracting}, which encourage representations to retain information about the input. More recently, masked autoencoders have achieved strong performance by reconstructing missing input regions \citep{he2022masked,bao2022beit}. Reconstruction naturally discourages collapse and can be interpreted as promoting high mutual information between inputs and their representations. However, input-space objectives may prioritize low-level reconstruction over downstream semantics \citep{pmlr-v235-balestriero24b}. \citet{chen2024deconstructing} connect denoising diffusion models to denoising autoencoders and propose \(\ell\)-DAE, which corrupts low-dimensional latent features but reconstructs the original input in pixel space using a ViT-based decoder. In contrast, EBReg operates entirely in representation space, using a score function to regularize latent representations directly. This avoids pixel-level reconstruction and decouples collapse prevention from input-space fidelity.

\section{Methodology}
\label{sec:method}

Our goal is to show that score matching itself can provide a viable objective for learning strong self-supervised representations. To this end, we first derive a mutual-information lower bound and introduce the score-based EBReg surrogate for optimizing marginal entropy (Section~\ref{sec:objective}). We then describe alternating score fitting, encoder learning, and score-guided test-time adaptation (Section~\ref{sec:ebreg-algorithm}), with an overview in Figure~\ref{fig:ebreg-overview} and training pseudocode in Appendix~\ref{app:ebreg-training}. Finally, we theoretically characterize the exact-score dynamics in collapsed and low-variance directions (Section~\ref{sec:ebreg-analysis}).

\begin{figure}[t]
    \centering
    \includegraphics[width=\linewidth]{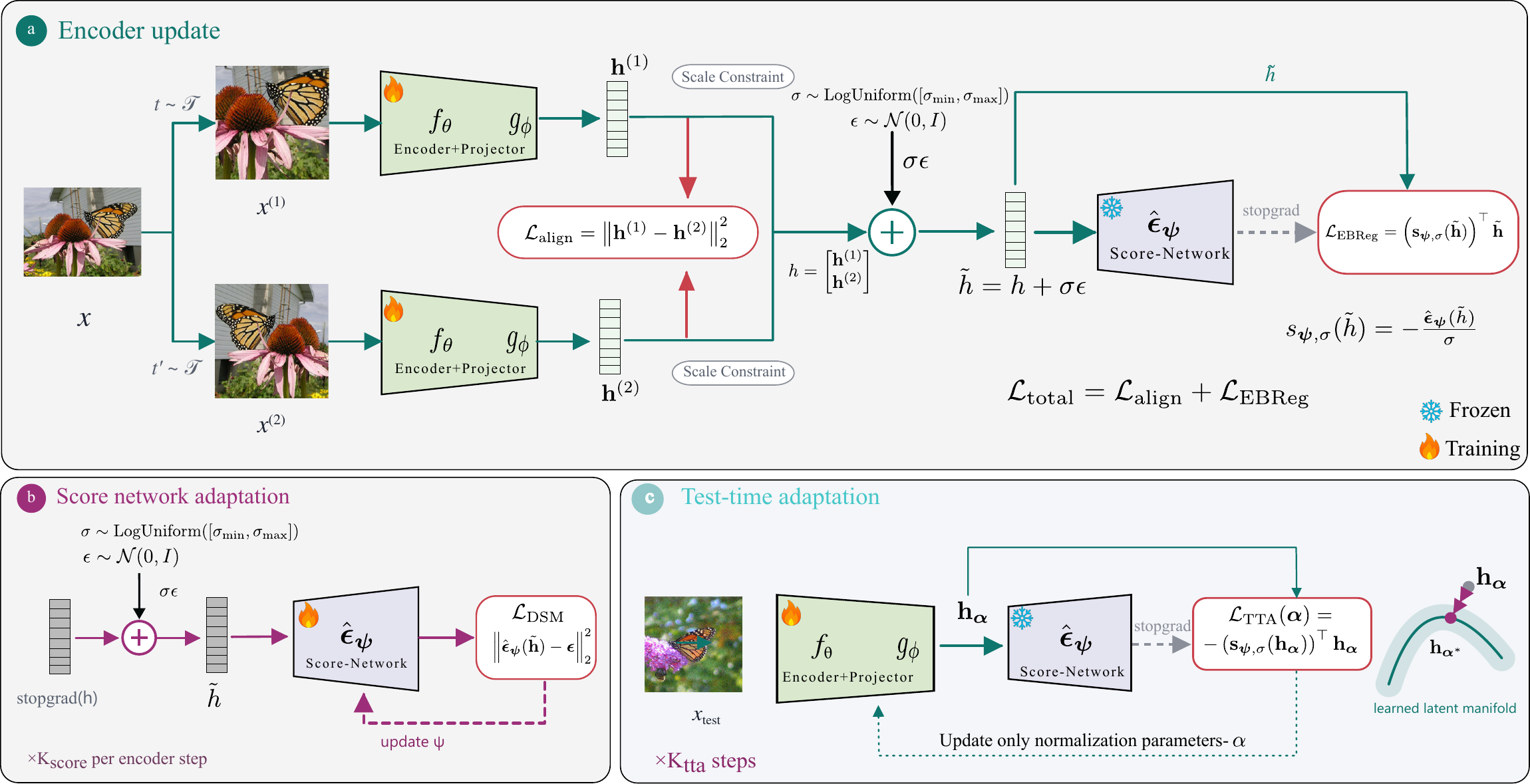}
    \caption{\textbf{Overview of EBReg: training and test-time adaptation.}
    \textbf{(a)} Two augmented views are encoded into normalized embeddings. The encoder minimizes a view-alignment loss and an entropy regularizer, $\mathcal{L}_{\mathrm{EBReg}}$, using a detached estimate of the score of the noise-smoothed embedding distribution.
    \textbf{(b)}. The latent score is parameterized by a noise-prediction network fitted on detached embeddings, with $K_{\mathrm{score}}$ updates per encoder step; it predicts the injected Gaussian noise, giving the score estimate $\vs_{\vpsi,\sigma}(\tilde \vh)=-\hat{\vepsilon}_\vpsi(\tilde \vh)/\sigma$.
    \textbf{(c)} At test time, the learned score is reused as a test-time adaptation signal. The score model and remaining encoder--projector parameters are frozen, while only normalization parameters $\valpha$ are updated by minimizing $\mathcal{L}_{\mathrm{TTA}}$, guiding corrupted embeddings toward higher-density regions of the learned source distribution.}
    \label{fig:ebreg-overview}
\end{figure}

\subsection{From mutual information to score-based entropy regularization}
\label{sec:objective}

EBReg combines alignment between augmented views with maximization of smoothed marginal entropy. We motivate this objective through the mutual information between representations of two augmented views. Maximizing mutual information favors representations that remain predictive across views while retaining variability in their marginal distribution. This leads naturally to an alignment--entropy formulation, where alignment promotes cross-view agreement and marginal entropy preserves representation diversity and prevents collapse. EBReg then uses a learned score to optimize the smoothed marginal entropy without explicit density evaluation.

\myparagraph{An alignment--entropy lower bound.}
Given an input $\rvx$, we draw two transformations $t,t'$ independently from an augmentation distribution $\mathcal{T}$. A shared encoder $f_\vtheta$ and projector $g_\vphi$ produce embeddings $\rvh^{(1)}=g_\vphi(f_\vtheta(t(\rvx)))$ and $\rvh^{(2)}=g_\vphi(f_\vtheta(t'(\rvx)))$ in $\R^d$. Both views share the marginal distribution $p_{\vtheta,\vphi}$. Embeddings may occupy a lower-dimensional support and lack a density in the full embedding space, making their differential entropy ill-defined. Gaussian smoothing yields a full-dimensional density, allowing us to work with a well-defined smoothed entropy. For a fixed noise scale $\sigma>0$, we define
\begin{equation}
    \tilde\rvh^{(1)}=\rvh^{(1)}+\sigma\rvepsilon,
    \qquad \rvepsilon\sim\mathcal{N}(\vzero,\mI_d),
    \label{eq:smoothed-entropy}
\end{equation}
where $\rvepsilon$ is independent of both views. We write $H_\sigma(\rvh)=H(\rvh+\sigma\rvepsilon)$, where $H$ denotes differential entropy, and $p_\sigma=p_{\vtheta,\vphi}*\mathcal{N}(\vzero,\sigma^2\mI_d)$ for the smoothed density.

To relate the smoothed representation to the mutual-information objective between the two clean views, we consider $I(\tilde{\rvh}^{(1)};\rvh^{(2)})$. Since $\tilde{\rvh}^{(1)}$ is obtained from $\rvh^{(1)}$ by adding independent Gaussian noise, $\rvh^{(2)}\to\rvh^{(1)}\to\tilde{\rvh}^{(1)}$ forms a Markov chain. The data processing inequality therefore gives $I(\rvh^{(1)};\rvh^{(2)})\geq I(\tilde{\rvh}^{(1)};\rvh^{(2)})$. Combining this inequality with the variational bound of \citet{barber2003im} applied to the smoothed pair yields
\begin{equation}
    I(\rvh^{(1)};\rvh^{(2)})\geq
    H_\sigma(\rvh^{(1)})+\E\log q(\tilde\rvh^{(1)}\mid\rvh^{(2)}),
    \label{eq:smoothed-mi-bound}
\end{equation}
for any variational conditional density $q$. We choose $q_\tau(\cdot\mid\rvh^{(2)})=\mathcal{N}(\rvh^{(2)},\tau^2\mI_d)$ with fixed $\tau>0$, yielding alignment loss. Choosing a Gaussian variational density does not require either the embedding distribution or the true conditional distribution to be Gaussian. Its cross-entropy is
\begin{equation}
    -\E\log q_\tau(\tilde\rvh^{(1)}\mid\rvh^{(2)})
    =\frac{d}{2\tau^2}\mathcal{L}_{\mathrm{align}}+C_{\sigma,\tau},
    \label{eq:gaussian-cross-entropy}
\end{equation}
where $\mathcal{L}_{\mathrm{align}}=d^{-1}\E\|\rvh^{(1)}-\rvh^{(2)}\|_2^2$ is the normalized alignment loss \citep{wang2020understanding} and $C_{\sigma,\tau}=\frac{d}{2}\log(2\pi\tau^2)+\frac{d\sigma^2}{2\tau^2}$ is independent of $(\vtheta,\vphi)$. See Appendix~\ref{app:mi-bound} for the full derivation. Substituting this cross-entropy expression into \Eqref{eq:smoothed-mi-bound} yields the alignment--entropy lower bound
\begin{equation}
    \boxed{I(\rvh^{(1)};\rvh^{(2)})
    \geq H_\sigma(\rvh)-\frac{d}{2\tau^2}\mathcal{L}_{\mathrm{align}}-C_{\sigma,\tau}.}
    \label{eq:ebreg-mi-lower-bound}
\end{equation}
Here, $\rvh$ denotes an embedding drawn from the shared marginal distribution of the two views. For fixed $\sigma$ and $\tau$, maximizing this lower bound is equivalent to minimizing $\mathcal{L}_{\mathrm{align}}-\gamma H_\sigma(\rvh)$, where $\gamma=2\tau^2/d$. The objective therefore balances agreement between paired views with entropy of their shared marginal distribution. Increasing latent scale can increase entropy without enriching the representation. We therefore constrain latent scale to prevent entropy maximization through rescaling alone, with the choice of constraint discussed in Section~\ref{sec:maxent-connections}.

\smallskip
\noindent
We now introduce a tractable gradient surrogate for optimizing the lower bound in \Eqref{eq:ebreg-mi-lower-bound}. The alignment term is computed directly from paired embeddings, but the smoothed entropy depends on the unknown density $p_\sigma$, which is generally intractable to evaluate. Rather than evaluating $p_\sigma$ directly, EBReg estimates its score, which points toward locally higher-density regions. We therefore replace the negative entropy term with the EBReg gradient surrogate and train the encoder on
\begin{equation}
    \boxed{\mathcal{L}_{\mathrm{total}}
    =\mathcal{L}_{\mathrm{align}}
    +\gamma\,\underbrace{\E\left[\operatorname{stopgrad}\left(\vs_{\vpsi,\sigma}(\tilde\rvh)\right)^\top\tilde\rvh\right]}_{\mathcal{L}_{\mathrm{EBReg}}}.}
    \label{eq:ebreg-objective}
\end{equation}
Here, $\tilde\rvh=\rvh+\sigma\rvepsilon$ and $\vs_{\vpsi,\sigma}$ denotes the score estimate. The stop-gradient treats the score as fixed when differentiating the surrogate with respect to the encoder and projector. Section~\ref{sec:ebreg-algorithm} describes the alternating score-fitting and encoder updates.

We now show that, with an exact score, this surrogate implements entropy maximization at the gradient level. The entropy-gradient identity for implicit distributions
\citep{li2018gradient,shi2018spectral} gives, under standard regularity conditions,
\begin{equation}
    \nabla_{\vtheta,\vphi} H_\sigma(\rvh)
    =-\E\left[\mJ_{\vtheta,\vphi}(\rvh)^\top \vs_\sigma(\tilde\rvh)\right],
    \label{eq:entropy-gradient}
\end{equation}
where $\mJ_{\vtheta,\vphi}(\rvh)=\partial\rvh/\partial(\vtheta,\vphi)$ is the embedding Jacobian. Detaching the entire score vector in \Eqref{eq:ebreg-objective} gives $\nabla_{\vtheta,\vphi}\mathcal{L}_{\mathrm{EBReg}}=\E[\mJ_{\vtheta,\vphi}(\rvh)^\top\vs_{\vpsi,\sigma}(\tilde\rvh)]$. Thus, with an exact score, the EBReg update implements descent on $-H_\sigma$ without evaluating the density or entropy. For $\gamma=2\tau^2/d$, the resulting representation-parameter gradient equals that of the negative MI lower bound up to the positive factor $\gamma$. With a learned score, this equivalence is approximate and the surrogate's value is neither an entropy estimate nor an MI bound.

Intuitively, alignment pulls paired embeddings together, while EBReg encourages the embedding distribution to spread out. The learned score \(\vs_{\vpsi,\sigma}(\tilde\rvh)\) approximates the local direction of increasing density in the noise-smoothed embedding distribution. Minimizing the surrogate encourages embeddings to move against the score, toward locally lower-density regions, thereby producing an entropy-promoting encoder update.

\myparagraph{Connections to existing regularizers.}
\label{sec:maxent-connections}
EBReg's scale constraint determines its entropy-maximizing target. Under zero mean and unit marginal variances, the maximizer is the isotropic Gaussian targeted by SIGReg~\citep{balestriero2025lejepaprovablescalableselfsupervised}, whose moments coincide with the variance--covariance conditions encouraged by VICReg ~\citep{bardes2022vicreg}.  More generally, different scale constraints induce different maximum-entropy targets while leaving the EBReg formulation unchanged. Table~\ref{tab:maxent-targets} relates these targets to existing SSL regularizers. However, sharing the same target does not imply equivalence of the corresponding objectives. Appendix~\ref{app:maxent-connections} provides the derivations.

\begin{table}[ht]
    \centering
    \caption{\textbf{Maximum-entropy targets under different latent constraints.}
    Each constraint induces a corresponding maximum-entropy target linked to existing SSL regularizers. First three targets maximize the smoothed entropy for any $\sigma>0$. The last holds as $\sigma\to0$
    (Appendix~\ref{app:maxent-connections}).}
    \label{tab:maxent-targets}
    \begingroup
    \small
    \setlength{\tabcolsep}{4pt}
    \renewcommand{\arraystretch}{1.05}
    \begin{tabular}{@{}p{0.23\linewidth}p{0.23\linewidth}p{0.20\linewidth}p{0.27\linewidth}@{}}
        \toprule
        \textbf{Latent constraint}
        & \textbf{Typical realization}
        & \textbf{Max.-entropy target}
        & \textbf{SSL connection} \\
        \midrule

        \raggedright $\E[\rvh]=\vzero$, $\operatorname{tr}\Cov(\rvh)=d$
        &
        \raggedright
        Batch normalization without affine parameters
        &
        \raggedright
        Isotropic Gaussian
        &
        \raggedright
        SIGReg: Gaussian target;\newline
        VICReg: corresponding first- and second-order moments
        \tabularnewline

        \addlinespace[2pt]

        \raggedright
        $\|\rvh\|_2=r$
        &
        \raggedright
        $\ell_2$, RMS, or layer normalization
        &
        \raggedright
        Uniform on the sphere
        &
        \raggedright
        Contrastive learning
        \tabularnewline

        \addlinespace[2pt]

        \raggedright
        $\rvh\in\{\ve_1,\ldots,\ve_K\}$
        &
        \raggedright
        Discrete prototype / code assignment
        &
        \raggedright
        Uniform code usage
        &
        \raggedright
        SwAV and MSN: balanced prototype usage;
        \tabularnewline

        \addlinespace[2pt]

        \raggedright
        $\E\|\rvh\|_p^p=M,\;
        \rvh\in S$
        &
        \raggedright
        $p$-th moment scaling with support constraint
        &
        \raggedright
        Generalized Gaussian
        (truncated to $S$)
        &
        \raggedright
        Rectified LpJEPA
        \tabularnewline

        \bottomrule
    \end{tabular}
    \endgroup
\end{table}

\subsection{Alternating score fitting and encoder learning}
\label{sec:ebreg-algorithm}

Since the score in \Eqref{eq:ebreg-objective} depends on the encoder's evolving embedding distribution, EBReg alternates between fitting the score model to the current embeddings and updating the encoder with the total objective. Figure~\ref{fig:ebreg-overview} summarizes these two phases. Algorithm~\ref{alg:ebreg-training} in Appendix~\ref{app:ebreg-training} provides the complete training procedure. Embedding normalization enforces the chosen scale constraint before Gaussian noise is added.

\myparagraph{Fitting the latent score.}
We parameterize a noise-prediction network $\hat{\vepsilon}_\vpsi$ and recover the latent score from its output via Tweedie's formula. For each minibatch, we pool embeddings from both augmented views and stop gradients through them, $\bar\rvh=\operatorname{stopgrad}(\rvh)$. With the encoder and projector held fixed, we take $K_{\mathrm{score}}$ updates of $\vpsi$ on the denoising objective
\begin{equation}
    \mathcal{L}_{\mathrm{DSM}}(\vpsi)
    =\E_{\bar\rvh,\rvepsilon}\left[\|\hat{\vepsilon}_\vpsi(\bar\rvh+\sigma\rvepsilon)-\rvepsilon\|_2^2\right],
    \qquad
    \vs_{\vpsi,\sigma}(\vz)=-\frac{\hat{\vepsilon}_\vpsi(\vz)}{\sigma},
    \label{eq:score-fitting}
\end{equation}
using fresh $\rvepsilon\sim\mathcal{N}(\vzero,\mI_d)$ for each update and the same noise level $\sigma$ as in Section~\ref{sec:objective}. At the unrestricted population optimum, the predictor equals $\E[\rvepsilon\mid\rvh+\sigma\rvepsilon=\vz]$, whose negative, scaled by $1/\sigma$ is the score of $p_\sigma$ \citep{vincent2011connection}. Finite network capacity and a finite number of updates yield an approximation to this score as the embedding distribution evolves.

\myparagraph{Updating the encoder.}
We recompute embeddings for the same paired views with gradients enabled, draw fresh noise, and update $(\vtheta,\vphi)$ using \Eqref{eq:ebreg-objective}. The alignment loss acts on clean paired embeddings while the entropy regularizer uses the score evaluated at their noisy counterparts. We detach the entire score output so that its value supplies the update direction while gradients flow through $\tilde\rvh$ to the encoder and projector. The score is fixed during the encoder update, so score fitting updates only $\vpsi$, and the encoder step updates only $(\vtheta,\vphi)$.

\subsection{Score-guided test-time adaptation}
\label{sec:score-tta}

Beyond its role in EBReg training, the learned score provides the direction of increasing log-density under the source latent distribution. This makes it directly useful for test-time adaptation, since corrupted inputs may produce representations that move away from high-density source regions and the score provides a direction back toward them. Let $\valpha$ denote the encoder's normalization parameters and $\rvh_{\valpha}$ the resulting test representation. We keep the score model, classifier, and remaining encoder parameters fixed, and adapt only $\valpha$, following ~\citep{wang2021tent}.

We define the score-guided adaptation objective as
\begin{equation}
    \mathcal{L}_{\mathrm{TTA}}(\valpha)
    =
    -\E\left[
    \operatorname{stopgrad}\left(
    \vs_{\vpsi,\sigma}(\rvh_{\valpha})
    \right)^\top
    \rvh_{\valpha}
    \right].
    \label{eq:tta-objective}
\end{equation}
The normalization parameters are updated for $K_{\mathrm{tta}}$ steps using gradients of
$\mathcal{L}_{\mathrm{TTA}}$, while the score model remains fixed. This objective reverses the role of the score used during EBReg training. During pretraining, the negative-score contribution promotes entropy by spreading representations away from high-density regions. At test time, the positive-score direction instead encourages corrupted representations toward higher-density regions of the source latent distribution. Thus, the score learned for representation regularization also provides a label-free signal for adaptation under distribution shift. Section~\ref{sec:exp-tta} evaluates this procedure under input corruptions.

\subsection{Collapse Dynamics and Connection to SSL Regularization}
\label{sec:ebreg-analysis}

We analyze EBReg near representational collapse and characterize how latent scale constraints determine maximum-entropy solutions related to existing SSL regularizers.

\myparagraph{Instability under dimensional collapse.}
If the embeddings lie in a lower-dimensional subspace of $\R^d$, evaluating the smoothed score on the clean embeddings yields no component along collapsed directions and therefore cannot recover them. Evaluating the score at noise-perturbed embeddings introduces a nonzero component in these directions. The following theorem formalizes
this behavior and shows that directions with small but nonzero variance are amplified.

\myparagraph{Intuition.}
Part~(a) explains why the score in \Eqref{eq:ebreg-objective} is evaluated at $\tilde\rvh$ rather than at $\rvh$. At a clean embedding, the score vanishes along any collapsed direction, regardless of the distribution within the occupied subspace. It also shows that exact collapse is a stationary point of the EBReg objective, with nonzero sample-wise updates along collapsed directions but zero mean drift.
Part~(b) shows that this stationary point is unstable, since the right-hand side of \Eqref{eq:collapsed-amplification} defines a variance-ascent direction and descending $\mathcal{L}_{\mathrm{total}}$ therefore amplifies any nonzero variance along $\vq$. Because the update scales with the variance already present rather than introducing a fixed increment, the relative amplification in the low-variance regime is governed by $\gamma/\sigma^2$ (Appendix~\ref{app:gaussian-ladder}). It is also not opposed by the alignment term, which penalizes disagreement between views while remaining invariant to variation shared across paired views. This theorem characterizes directional gradient contributions under an exact score and, in part (b), an independent Gaussian-coordinate assumption; it does not guarantee recovery from exact collapse or variance growth under the full training dynamics.
\begin{theorem}[Instability of collapsed directions under EBReg]
\label{thm:collapsed-signal}
Let $\rvh\in\R^d$ have finite second moment and fix a unit direction $\vq$. Assume an exact score $\vs_\sigma$ and the Gaussian perturbation above, with noise independent of the input and augmentation, and write $\rva_\vq=\mJ_{\vtheta,\vphi}(\rvh)^\top\vq$ for the parameter gradient of $\vq^\top\rvh$, with $\E\|\rva_\vq\|_2^2<\infty$.

\textbf{(a) The clean score is blind to dimensional collapse.} If $\operatorname{Var}(\vq^\top\rvh)=0$, then almost surely
\begin{equation}
    \vq^\top\vs_\sigma(\tilde\rvh)=-\frac{\vq^\top\rvepsilon}{\sigma},
    \qquad
    \vq^\top\vs_\sigma(\rvh)=0,
    \label{eq:collapsed-score}
\end{equation}
so the per-sample gradient through $\vq$, $\rvg_\vq=-(\gamma/\sigma)(\vq^\top\rvepsilon)\,\rva_\vq$, is nonzero but has zero mean.

\textbf{(b) Amplification.} If $\vq^\top\rvh=c+\delta\ru$ with $\delta>0$ and $\ru\sim\mathcal{N}(0,1)$ independent of $(\mI_d-\vq\vq^\top)\rvh$, then
\begin{equation}
    \gamma\,\E\bigl[\rva_\vq\,\vq^\top\vs_\sigma(\tilde\rvh)\bigr]
    =-\frac{\gamma}{2(\delta^2+\sigma^2)}\,\nabla_{\vtheta,\vphi}\operatorname{Var}(\vq^\top\rvh).
    \label{eq:collapsed-amplification}
\end{equation}
\end{theorem}

\section{Experiments}
\label{sec:experiments}
We compare EBReg with existing self-supervised methods, matching parameter count, data augmentations, and training epochs unless otherwise specified. We evaluate representation quality, collapse prevention, hyperparameter sensitivity, computational overhead, and test-time adaptation under input corruptions.

\providecommand{\pending}[1]{\textbf{[#1]}}
\providecommand{\tbd}{\textbf{?}}

\subsection{Linear evaluation}
\label{sec:exp-linear}
We evaluate the effectiveness of EBReg across multiple architectures and self-supervised objectives on ImageNet-1K \citep{imagenet}. We consider ResNet-50 \citep{He_2016_CVPR}, ViT-Large, and ViT-Huge \citep{dosovitskiy2021an} under three established pretraining protocols. For ResNet-50, we adopt the setup from \citet{bardes2022vicreg}, including its data augmentations, optimization settings, and training schedule, and replace the original variance--covariance regularization with EBReg. For ViT-Large, we follow the setup from \citet{balestriero2025lejepaprovablescalableselfsupervised} and replace SIGReg with EBReg while keeping the remaining pretraining and evaluation protocol unchanged. Finally, we evaluate EBReg with ViT-Huge in the setup from \citet{Assran_2023_CVPR}, where EBReg is added to the existing objective rather than replacing it. This setting assesses whether score-based entropy regularization can provide a complementary signal to an existing self-supervised objective. Within each setting, EBReg and the corresponding baseline use the same backbone, augmentations, training schedule, and evaluation procedure. The results are summarized in Table~\ref{tab:linear}.

\begin{table}[t]
    \centering
    \caption{
    \textbf{Linear-probe performance and effective rank on ImageNet-1K.}
    Results are grouped by pretraining setup. ResNet-50, ViT-L, and ViT-H use the setups from
    \citet{bardes2022vicreg}, \citet{balestriero2025lejepaprovablescalableselfsupervised},
    and \citet{Assran_2023_CVPR}, respectively. Within each setup, EBReg and the
    corresponding baselines use the same backbone, data augmentations, optimization
    settings, training schedule, and evaluation procedure. Effective rank
    \citep{garrido2023rankme} is computed on backbone representations after
    pretraining, with higher values indicating less dimensional collapse. Bold indicates higher accuracy within each setup, and underline indicates the second-best.
    }
    \label{tab:linear}

  \begingroup
\setlength{\tabcolsep}{5pt}
\renewcommand{\arraystretch}{1.12}
\small
\begin{tabular*}{\linewidth}{@{\extracolsep{\fill}}llccc@{}}
    \toprule
    \textbf{Backbone}
    & \textbf{Method / Regularizer}
    & \textbf{Epochs}
    & \textbf{Linear probe (\%)} $\uparrow$
    & \textbf{Eff. rank} $\uparrow$ \\
    \midrule

    \multirow{4}{*}{ResNet-50}
        & Alignment only ($\gamma=0$) & 100 & Collapse & -- \\
        & VICReg & 100 & \textbf{68.02} & 1368 \\
        & SIGReg & 100 & 66.42 & 1256 \\
        & \textbf{EBReg (ours)} & 100 & \underline{67.92} & \textbf{1431} \\
    \midrule

    \multirow{3}{*}{ViT-L}
        & SIGReg & 100 & \underline{75.32} & 648 \\
        & $\ell$-DAE & 1600 & 75.00 & -- \\
        & \textbf{EBReg (ours)} & 100 & \textbf{75.57} & \textbf{759} \\
    \midrule

    \multirow{2}{*}{ViT-H}
        & I-JEPA & 100 & 79.03 & 540 \\
        & I-JEPA + StoP & 100 & \underline{79.33} & 527 \\
        & \textbf{I-JEPA + StoP + EBReg (ours)}
        & 100 & \textbf{79.41} & \textbf{957} \\
    \bottomrule
\end{tabular*}
\endgroup
\end{table}

Table~\ref{tab:linear} shows that EBReg remains competitive in linear evaluation while consistently producing higher-rank representations across architectures. On ResNet-50, EBReg matches the performance of strong regularization-based baselines while yielding the highest effective rank. On ViT-L, it improves both linear-probe accuracy and representation rank relative to SIGReg.When added to StoP on ViT-H, EBReg yields a modest improvement in linear-probe accuracy while substantially increasing effective rank, suggesting that it complements the existing objective while preserving a richer representation space.

\subsection{Hyperparameter Ablations and Computational Cost}
\label{sec:exp-sensitivity}

\myparagraph{Hyperparameter ablations.}
We  ablate hyperparameter under the setup from \citet{bardes2022vicreg} and report ImageNet-1K linear-probe accuracy in Table~\ref{tab:sensitivity}. Among the tested values, the default configuration $\gamma=0.34$, $\sigma=4.5$, and $K_{\mathrm{score}}=16$ yields the highest accuracy. Performance is most sensitive to the noise scale and the number of score updates, while projector dimension has comparatively little effect over the tested range. Increasing score-network width beyond the default also degrades performance, indicating that greater score-model capacity or more frequent score fitting does not necessarily improve representation quality.

\myparagraph{Computational cost.}
Table~\ref{tab:cost} reports the computational cost of EBReg under the setup from \citet{balestriero2025lejepaprovablescalableselfsupervised} using identical hardware. For a backbone with approximately $300$M parameters, the score network adds only $5.21$M additional parameters, under $2\%$ of the backbone parameter count. The primary overhead arises from repeatedly updating the score model to track the evolving latent distribution. Relative to SIGReg, this increases wall-clock training time by approximately $60$ seconds per epoch, from $440$ to $500$ seconds, corresponding to a $13.6\%$ increase.

The additional computation remains moderate because score fitting is performed entirely in the projected latent space. Its cost is therefore determined primarily by the latent dimension, score-network architecture, and number of score updates, rather than by the image resolution or backbone size. The strong performance obtained with a compact score network may indicate that the latent distribution admits a sufficiently structured score field that can be approximated with limited capacity, although we do not directly characterize its intrinsic dimensionality.

\begin{table}[ht]
    \centering
    \begin{minipage}[t]{0.57\linewidth}
        \centering
        \captionof{table}{ResNet-50 sensitivity: ImageNet-1K linear-probe accuracy, varying one factor at a time. Defaults are \underline{underlined}.}
        \label{tab:sensitivity}
        \scriptsize
        \renewcommand{\arraystretch}{1.15}
        \resizebox{\linewidth}{!}{%
            \begin{tabular}{@{}llccc@{}}
                \hline
                \textbf{Factor} & \textbf{Values} & \multicolumn{3}{c}{\textbf{Accuracy (\%)}} \\
                \hline
                Regularization weight $\gamma$ & 0.25 / \underline{0.34} / 0.5 & 64.28 & 67.92 & 63.38 \\
                Noise level $\sigma$ & 4.25 / \underline{4.5} / 5.0 & 62.32 & 67.92 & 65.43 \\
                Score updates $K_{\mathrm{score}}$ & 4 / \underline{16} / 32 & 63.72 & 67.92 & 58.32 \\
                Score-network width & 1024 / \underline{2048} / 4096 & 67.65 & 67.92 & 61.35 \\
                $\sigma$ width  & 0.1 / 0.5 / 0.7 & 67.51 & 67.92 & 67.43 \\
                Projector dimension $d$ & 64 / \underline{256} / 512 & 67.42 & 67.92 & 67.51 \\
                \hline
            \end{tabular}%
        }
    \end{minipage}\hfill
    \begin{minipage}[t]{0.42\linewidth}
        \centering
        \captionof{table}{ViT-L training costs on identical hardware. Extra parameters are relative to the backbone and projector.}
        \label{tab:cost}
        \footnotesize
        \setlength{\tabcolsep}{3pt}
        \renewcommand{\arraystretch}{1.15}
        \begin{tabular}{@{}lrr@{}}
            \toprule
            \textbf{Metric} & \textbf{SIGReg} & \textbf{EBReg} \\
            \midrule
            Extra params (M) & 0 & 5.21 \\
            Time / epoch (s) & 440 & 500 \\
            Peak memory (GB) & 90.33 & 89.84 \\
            GPU-hours & 195.83 & 224.01 \\
            \bottomrule
        \end{tabular}
    \end{minipage}
\end{table}

\begin{table}[ht]
    \centering
    \caption{ImageNet-C accuracy (\%) before and after five adaptation steps.
    Seven corruptions with gains for every model are selected by mean gain across models;
    the final row averages all 15 corruptions.
    Bold indicates higher accuracy within each model.}
    \label{tab:tta}

    \begingroup
    \small
    \setlength{\tabcolsep}{3pt}
    \renewcommand{\arraystretch}{1.12}

    \begin{tabular*}{\linewidth}{@{\extracolsep{\fill}}
        l
        rr
        rr
        rr
        @{}}
        \toprule
        &
        \multicolumn{2}{c}{
            \parbox{0.20\linewidth}{\centering
                \textbf{ViT-L}\\[-1pt]
                \scriptsize Setup from
                \citet{balestriero2025lejepaprovablescalableselfsupervised}
            }}
        &
        \multicolumn{2}{c}{
            \parbox{0.20\linewidth}{\centering
                \textbf{ViT-H}\\[-1pt]
                \scriptsize Setup from
                \citet{Assran_2023_CVPR}
            }}
        &
        \multicolumn{2}{c}{
            \parbox{0.20\linewidth}{\centering
                \textbf{ResNet-50}\\[-1pt]
                \scriptsize Setup from
                \citet{bardes2022vicreg}
            }}
        \\
        \cmidrule(lr){2-3}
        \cmidrule(lr){4-5}
        \cmidrule(lr){6-7}

        \textbf{Corruption}
        & \textbf{Before} & \textbf{After}
        & \textbf{Before} & \textbf{After}
        & \textbf{Before} & \textbf{After} \\
        \midrule

        Fog
        & 44.18 & \textbf{46.34}
        & 58.41 & \textbf{61.42}
        & 24.00 & \textbf{38.60} \\

        Impulse noise
        & 43.41 & \textbf{46.86}
        & 50.77 & \textbf{56.33}
        & 15.52 & \textbf{22.71} \\

        Motion blur
        & 41.12 & \textbf{46.29}
        & 58.01 & \textbf{61.52}
        & 18.86 & \textbf{25.37} \\

        Glass blur
        & 21.72 & \textbf{25.39}
        & 44.75 & \textbf{49.93}
        & 7.78 & \textbf{13.59} \\

        Zoom blur
        & 39.26 & \textbf{43.00}
        & 50.92 & \textbf{54.70}
        & 20.19 & \textbf{27.27} \\

        Gaussian noise
        & 44.56 & \textbf{48.21}
        & 49.17 & \textbf{55.97}
        & 25.78 & \textbf{25.99} \\

        Contrast
        & 57.79 & \textbf{58.50}
        & 60.44 & \textbf{63.70}
        & 40.94 & \textbf{44.81} \\

        \midrule
        Mean (15 corruptions)
        & 46.55 & \textbf{48.18}
        & 57.79 & \textbf{61.05}
        & 28.67 & \textbf{30.50} \\

        \bottomrule
    \end{tabular*}
    \endgroup
\end{table}

\subsection{Score-guided test-time adaptation}
\label{sec:exp-tta}
We evaluate score-guided test-time adaptation on ImageNet-C~\citep{hendrycks2019benchmarking} using the EBReg-trained models and pretraining setups from Section~\ref{sec:exp-linear}. In each case, EBReg training yields a corresponding latent score model, which we reuse at test time as the adaptation signal. The score model and classifier remain fixed, while only the encoder's normalization parameters are updated using the objective from Section~\ref{sec:score-tta}. Table~\ref{tab:tta} reports accuracy before and after five adaptation steps.

Score-guided adaptation improves mean accuracy across all three EBReg-trained models and a broad range of corruptions. These gains indicate that the learned score remains informative under distribution shift, guiding corrupted representations toward higher-density regions of the corresponding source latent distribution. Improvements are particularly pronounced for several noise and blur corruptions, though their magnitude varies across architectures and corruption types.

\section{Conclusion}
\label{sec:conclusion}
In this paper, we introduce EBReg, a representation learning framework derived from a lower bound on mutual information. EBReg jointly trains the encoder with a lightweight score network and uses the learned score to counter representation collapse. Our analysis shows that dimension-collapsed solutions are unstable under the proposed objective, which induces a persistent force toward activating collapsed dimensions. Under constraints preventing unbounded feature scaling, EBReg recovers target structures related to existing self-supervised regularizers. Empirically, EBReg consistently learns informative, non-collapsed representations across a range of settings. At test time, the learned score guides corrupted representations toward the learned manifold, improving robustness to noise.

\section*{Ethics Statement}
This work addresses fundamental research questions in self-supervised representation learning and entropy-based regularization. We do not identify ethical concerns specific to the proposed method or the experiments reported in this paper. As with other representation-learning methods, downstream applications should be assessed for potential bias, privacy implications, and misuse in their intended context.

\section*{Reproducibility Statement}
Section~\ref{sec:method} specifies the training objective, alternating optimization procedure, and test-time adaptation rule, with pseudocode in Appendix~\ref{app:ebreg-training}. Section~\ref{sec:experiments} describes the datasets, backbone architectures, baseline protocols, and evaluation settings; Tables~\ref{tab:sensitivity} and~\ref{tab:cost} report hyperparameter ablations and computational costs. The derivation of the mutual-information bound, analysis of collapsed directions, and maximum-entropy connections are provided in Appendices~\ref{app:mi-bound}, \ref{app:gaussian-ladder}, and~\ref{app:maxent-connections}, respectively.

\bibliography{iclr2027_conference}
\bibliographystyle{iclr2027_conference}

\appendix
\section{The smoothed variational bound}
\label{app:mi-bound}

For completeness we record the two steps between the data processing inequality and \Eqref{eq:smoothed-mi-bound}; both are the standard variational decomposition of mutual information \citep{barber2003im,poole2019variational}, applied to the smoothed variable $\tilde\rvh^{(1)}=\rvh^{(1)}+\sigma\rvepsilon$. Writing the conditional entropy as a cross-entropy and then inserting $p=(p/q)\,q$,
{\allowdisplaybreaks
\begin{align}
    I(\tilde\rvh^{(1)};\rvh^{(2)})
    &=H_\sigma(\rvh^{(1)})+\E_{\tilde\rvh^{(1)},\rvh^{(2)}}\log p(\tilde\rvh^{(1)}\mid\rvh^{(2)})\nonumber\\
    &=H_\sigma(\rvh^{(1)})+\E_{\tilde\rvh^{(1)},\rvh^{(2)}}\log q(\tilde\rvh^{(1)}\mid\rvh^{(2)})
      +\underbrace{\E_{\rvh^{(2)}}D_{\mathrm{KL}}\bigl(p(\cdot\mid\rvh^{(2)})\,\|\,q(\cdot\mid\rvh^{(2)})\bigr)}_{\geq 0}\nonumber\\
    &\geq H_\sigma(\rvh^{(1)})+\E_{\tilde\rvh^{(1)},\rvh^{(2)}}\log q(\tilde\rvh^{(1)}\mid\rvh^{(2)}),
    \label{eq:smoothed-mi-bound-appendix}
\end{align}}%
where $q(\cdot\mid\vh^{(2)})$ is any conditional density. Both differential entropies are finite because $\tilde\rvh^{(1)}$ has a density for every $\sigma>0$ and the embeddings have finite second moments, so the decomposition is valid even when $\rvh^{(1)}$ itself is supported on a set of measure zero.

\paragraph{Gaussian variational conditional.}
For $q_\tau(\cdot\mid\vh^{(2)})=\mathcal{N}(\vh^{(2)},\tau^2\mI_d)$, independence and the moments of the Gaussian noise give
\begin{align}
    &-\E_{\tilde\rvh^{(1)},\rvh^{(2)}}\log q_\tau(\tilde\rvh^{(1)}\mid\rvh^{(2)})
    =\frac{d}{2}\log(2\pi\tau^2)
      +\frac{1}{2\tau^2}\E\|\rvh^{(1)}+\sigma\rvepsilon-\rvh^{(2)}\|_2^2\nonumber\\
    &\quad=\frac{d}{2}\log(2\pi\tau^2)
      +\frac{1}{2\tau^2}\Bigl[\E\|\rvh^{(1)}-\rvh^{(2)}\|_2^2
      +\underbrace{2\sigma\,\E\bigl[(\rvh^{(1)}-\rvh^{(2)})^\top\rvepsilon\bigr]}_{=0}
      +\underbrace{\sigma^2\,\E\|\rvepsilon\|_2^2}_{=\sigma^2 d}\Bigr]\nonumber\\
    &\quad=\frac{1}{2\tau^2}\E\|\rvh^{(1)}-\rvh^{(2)}\|_2^2+C_{\sigma,\tau},
    \label{eq:gaussian-cross-entropy-appendix}
\end{align}
where $C_{\sigma,\tau}=\frac{d}{2}\log(2\pi\tau^2)+\frac{d\sigma^2}{2\tau^2}$. The gap between the original MI and the lower bound is the sum of the smoothing loss $I(\rvh^{(1)};\rvh^{(2)})-I(\tilde\rvh^{(1)};\rvh^{(2)})$ and the conditional KL divergence in \Eqref{eq:smoothed-mi-bound-appendix}. The variational step is tight when $q$ equals the true smoothed conditional density almost surely. Under finite second moments, the smoothed MI and the Gaussian conditional KL term are finite, whereas the smoothing loss can be infinite if the original MI is infinite.

\section{Dimensional collapse}
\label{app:gaussian-ladder}

\paragraph{Setup and notation.}
Let $\rvh\in\R^d$ be an embedding with mean $\vmu$, covariance $\mSigma=\Cov(\rvh)$, and finite second moment. Let $\rvepsilon\sim\mathcal{N}(\vzero,\mI_d)$ be independent of $\rvh$ and of the input and augmentation that generate it. The noisy embedding is $\tilde\rvh=\rvh+\sigma\rvepsilon$; when we want to stress that it is a random variable with a density we write $\rvz$ for it. Conditionally on $\rvh$, $\rvz$ is Gaussian with mean $\rvh$ and covariance $\sigma^2\mI_d$, so averaging over $\rvh$ gives its density
\begin{equation}
    p_\sigma(\vz)=c_\sigma\,\E_{\rvh}\Bigl[\exp\Bigl(-\frac{\|\vz-\rvh\|_2^2}{2\sigma^2}\Bigr)\Bigr],
    \qquad c_\sigma=(2\pi\sigma^2)^{-d/2}.
    \label{eq:smoothed-density}
\end{equation}
This density is smooth and strictly positive for every $\sigma>0$, even when $\rvh$ itself is discrete or lies on a lower-dimensional set. The score is $\vs_\sigma(\vz)=\nabla_\vz\log p_\sigma(\vz)$.

\subsection{Proof of Theorem~\ref{thm:collapsed-signal}}

EBReg approximates the gradient of $-H_\sigma$. The entropy objective itself disfavors dimensional collapse, since under a fixed variance budget its unique maximizer is full rank (Appendix~\ref{app:maxent-connections}, part~(i)). That objective-level fact does not describe what the EBReg update does to a collapsed encoder; Theorem~\ref{thm:collapsed-signal} does, and we prove it here.

\subsubsection*{Part (a): the clean score is blind}

\emph{Goal.} Assuming $\operatorname{Var}(\vq^\top\rvh)=0$, show that $\vq^\top\vs_\sigma(\rvh)=0$ and $\vq^\top\vs_\sigma(\tilde\rvh)=-\vq^\top\rvepsilon/\sigma$, and that the resulting gradient has zero mean. \emph{Idea.} If the embeddings lie in a subspace, then in the directions orthogonal to it the noisy embedding is pure Gaussian noise, and the score of a Gaussian is known in closed form.

\medskip\noindent\textbf{Step 1: the embeddings lie in a subspace.}
Let $S=\operatorname{range}(\mSigma)$ be the subspace the embeddings occupy, $r=\dim S$, $\mP$ the orthogonal projector onto $S$, and $\mP_\perp=\mI_d-\mP$ the projector onto $S^\perp$. The collapsed directions are exactly the vectors of $S^\perp$. For any $\vv\in S^\perp$,
\begin{equation*}
    \operatorname{Var}(\vv^\top\rvh)=\vv^\top\mSigma\vv=0
    \quad\Longrightarrow\quad
    \vv^\top(\rvh-\vmu)=0\ \text{almost surely}.
\end{equation*}
Applying this to a basis of $S^\perp$ gives $\mP_\perp(\rvh-\vmu)=\vzero$ almost surely: every embedding equals the mean plus a vector in $S$.

\medskip\noindent\textbf{Step 2: split the distance into two orthogonal pieces.}
Fix a point $\vz$. Insert $\vmu$ and split $\vz-\vmu$ into its components in $S$ and $S^\perp$:
\begin{equation*}
    \vz-\rvh=\underbrace{\bigl[\mP(\vz-\vmu)-(\rvh-\vmu)\bigr]}_{\in S\text{ by Step 1}}
    +\underbrace{\mP_\perp(\vz-\vmu)}_{\in S^\perp}.
\end{equation*}
The two pieces are orthogonal, so their squared norms add:
\begin{equation*}
    \|\vz-\rvh\|_2^2=\bigl\|\mP(\vz-\vmu)-(\rvh-\vmu)\bigr\|_2^2+\bigl\|\mP_\perp(\vz-\vmu)\bigr\|_2^2 .
\end{equation*}
Only the first term depends on $\rvh$.

\medskip\noindent\textbf{Step 3: the density factorizes.}
Substitute Step~2 into \Eqref{eq:smoothed-density}. The exponential of a sum is a product, and the factor that does not depend on $\rvh$ comes out of the expectation:
\begin{align*}
    p_\sigma(\vz)&=p_S\bigl(\mP(\vz-\vmu)\bigr)\times
    \underbrace{(2\pi\sigma^2)^{-(d-r)/2}\exp\Bigl(-\frac{\|\mP_\perp(\vz-\vmu)\|_2^2}{2\sigma^2}\Bigr)}_{\text{Gaussian in }S^\perp},\\
    p_S(\vy)&:=(2\pi\sigma^2)^{-r/2}\,\E_\rvh\Bigl[\exp\Bigl(-\frac{\|\vy-(\rvh-\vmu)\|_2^2}{2\sigma^2}\Bigr)\Bigr].
\end{align*}
The first factor depends on $\vz$ only through its component in $S$, and carries everything about the embedding distribution. The second is a Gaussian density in $S^\perp$ with variance $\sigma^2$, and is the same whatever the embeddings do inside $S$.

\medskip\noindent\textbf{Step 4: the score in the collapsed directions.}
Take the logarithm and differentiate:
\begin{equation*}
    \vs_\sigma(\vz)=\nabla_\vz\log p_S\bigl(\mP(\vz-\vmu)\bigr)-\frac{\mP_\perp(\vz-\vmu)}{\sigma^2}.
\end{equation*}
By the chain rule the first term equals $\mP\,\nabla\log p_S$, so it lies in $S$. Applying $\mP_\perp$ removes it, because $\mP_\perp\mP=\vzero$, and leaves
\begin{equation}
    \mP_\perp\vs_\sigma(\vz)=-\frac{\mP_\perp(\vz-\vmu)}{\sigma^2}\qquad\text{for every }\vz\in\R^d .
    \label{eq:perp-score}
\end{equation}

\medskip\noindent\textbf{Step 5: evaluate at clean and noisy embeddings.}
At a clean embedding, Step~1 gives $\mP_\perp(\rvh-\vmu)=\vzero$, so $\mP_\perp\vs_\sigma(\rvh)=\vzero$. At a noisy embedding,
\begin{equation*}
    \mP_\perp(\tilde\rvh-\vmu)=\underbrace{\mP_\perp(\rvh-\vmu)}_{=\vzero}+\sigma\mP_\perp\rvepsilon
    \quad\Longrightarrow\quad
    \mP_\perp\vs_\sigma(\tilde\rvh)=-\frac{\mP_\perp\rvepsilon}{\sigma}.
\end{equation*}
For a unit collapsed direction $\vq\in S^\perp$ we have $\vq^\top\mP_\perp=\vq^\top$, so taking the inner product with $\vq$ gives \Eqref{eq:collapsed-score}.

\medskip\noindent\textbf{Step 6: the gradient through $\vq$ has zero mean.}
The per-sample EBReg loss is $\gamma\operatorname{stopgrad}(\vs_\sigma(\tilde\rvh))^\top\tilde\rvh$. The score is detached and the noise does not depend on the parameters, so by the chain rule its parameter gradient is $\gamma\,\mJ_{\vtheta,\vphi}(\rvh)^\top\vs_\sigma(\tilde\rvh)$. The part of this gradient that passes through direction $\vq$ is
\begin{equation*}
    \rvg_\vq=\gamma\,\underbrace{\mJ_{\vtheta,\vphi}(\rvh)^\top\vq}_{=\,\rva_\vq}\;\underbrace{\vq^\top\vs_\sigma(\tilde\rvh)}_{=\,-\vq^\top\rvepsilon/\sigma}
    =-\frac{\gamma}{\sigma}\,(\vq^\top\rvepsilon)\,\rva_\vq .
\end{equation*}
The vector $\rva_\vq$ depends only on the view, while $\vq^\top\rvepsilon\sim\mathcal{N}(0,1)$ is independent of it. Therefore
\begin{equation*}
    \E[\rvg_\vq]=-\frac{\gamma}{\sigma}\,\E[\vq^\top\rvepsilon]\,\E[\rva_\vq]=\vzero,
    \qquad
    \E\|\rvg_\vq\|_2^2=\frac{\gamma^2}{\sigma^2}\,\E\bigl[(\vq^\top\rvepsilon)^2\bigr]\,\E\|\rva_\vq\|_2^2=\frac{\gamma^2}{\sigma^2}\,\E\|\rva_\vq\|_2^2 .
\end{equation*}
So each sample contributes a nonzero push along $\vq$, but the pushes average to zero: exact collapse is a stationary point of the EBReg term. \qed

\subsubsection*{Part (b): amplification}

\emph{Goal.} Assuming $\vq^\top\rvh=c+\delta\ru$ with $\delta>0$, where $\ru\sim\mathcal{N}(0,1)$ is independent of the remaining coordinates $(\mI_d-\vq\vq^\top)\rvh$, show that the expected EBReg gradient through $\vq$ is a negative multiple of $\nabla\operatorname{Var}(\vq^\top\rvh)$. \emph{Idea.} Along $\vq$ the noisy embedding is an independent one-dimensional Gaussian, whose score is linear; the rest is a covariance computation. Write $\Theta=(\vtheta,\vphi)$ for the parameters.

\medskip\noindent\textbf{Step 1: the noisy coordinate along $\vq$ is an independent Gaussian.}
Along $\vq$,
\begin{equation*}
    \vq^\top\rvz=c+\delta\ru+\sigma\,\vq^\top\rvepsilon .
\end{equation*}
This is a constant plus two independent Gaussians with variances $\delta^2$ and $\sigma^2$, so $\vq^\top\rvz\sim\mathcal{N}(c,\delta^2+\sigma^2)$. It is also independent of the remaining coordinates $(\mI_d-\vq\vq^\top)\rvz=(\mI_d-\vq\vq^\top)\rvh+\sigma(\mI_d-\vq\vq^\top)\rvepsilon$: $\ru$ is independent of the first term by assumption, and for isotropic Gaussian noise the component $\vq^\top\rvepsilon$ is independent of the components orthogonal to $\vq$.

\medskip\noindent\textbf{Step 2: the score along $\vq$.}
Independence means the density is a product of the one-dimensional Gaussian density of $\vq^\top\vz$ and a density of the remaining coordinates. Only the first depends on $\vq^\top\vz$, so
\begin{equation*}
    \vq^\top\vs_\sigma(\vz)=\frac{\partial}{\partial(\vq^\top\vz)}\Bigl[-\frac{(\vq^\top\vz-c)^2}{2(\delta^2+\sigma^2)}\Bigr]=-\frac{\vq^\top\vz-c}{\delta^2+\sigma^2}.
\end{equation*}
At $\vz=\tilde\rvh$ we have $\vq^\top\tilde\rvh-c=\delta\ru+\sigma\,\vq^\top\rvepsilon$, hence
\begin{equation*}
    \vq^\top\vs_\sigma(\tilde\rvh)=-\frac{\delta\ru+\sigma\,\vq^\top\rvepsilon}{\delta^2+\sigma^2}.
\end{equation*}
Setting $\delta=0$ recovers part~(a).

\medskip\noindent\textbf{Step 3: the expected gradient through $\vq$.}
Multiply by $\rva_\vq$ and take expectations:
\begin{equation*}
    \E\bigl[\rva_\vq\,\vq^\top\vs_\sigma(\tilde\rvh)\bigr]
    =-\frac{1}{\delta^2+\sigma^2}\Bigl(\delta\,\E[\rva_\vq\ru]+\sigma\underbrace{\E[\rva_\vq]\,\E[\vq^\top\rvepsilon]}_{=\vzero}\Bigr)
    =-\frac{\delta\,\Cov(\rva_\vq,\ru)}{\delta^2+\sigma^2}.
\end{equation*}
The noise term vanishes because $\rvepsilon$ is independent of the view and has zero mean, and $\E[\rva_\vq\ru]=\Cov(\rva_\vq,\ru)$ because $\E[\ru]=0$. The noise no longer cancels everything: the embedding's own variation $\delta\ru$ correlates with how the encoder responds.

\medskip\noindent\textbf{Step 4: the gradient of the variance.}
Differentiate $\operatorname{Var}(\vq^\top\rvh_\Theta)=\E\bigl[(\vq^\top\rvh_\Theta-\E[\vq^\top\rvh_\Theta])^2\bigr]$ under the expectation, which $\E\|\rva_\vq\|_2^2<\infty$ and local domination of the parameter derivative permit. Using $\nabla_\Theta(\vq^\top\rvh_\Theta)=\rva_\vq$,
\begin{equation*}
    \nabla_\Theta\operatorname{Var}(\vq^\top\rvh_\Theta)
    =2\,\E\bigl[(\vq^\top\rvh-\E[\vq^\top\rvh])(\rva_\vq-\E[\rva_\vq])\bigr]
    =2\Cov(\rva_\vq,\vq^\top\rvh)
    =2\delta\Cov(\rva_\vq,\ru),
\end{equation*}
where the last step uses $\vq^\top\rvh=c+\delta\ru$.

\medskip\noindent\textbf{Step 5: combine.}
Steps~3 and~4 contain the same covariance. Eliminating it,
\begin{equation*}
    \gamma\,\E\bigl[\rva_\vq\,\vq^\top\vs_\sigma(\tilde\rvh)\bigr]
    =-\frac{\gamma\,\delta\Cov(\rva_\vq,\ru)}{\delta^2+\sigma^2}
    =-\frac{\gamma}{2(\delta^2+\sigma^2)}\,\nabla_\Theta\operatorname{Var}(\vq^\top\rvh),
\end{equation*}
which is \Eqref{eq:collapsed-amplification}. \qed

\medskip\noindent\textbf{Consequence: exponential growth.}
Gradient descent subtracts $\eta$ times the gradient, so the contribution of this term to the update is $\Delta\Theta=\frac{\eta\gamma}{2(\delta^2+\sigma^2)}\nabla_\Theta\operatorname{Var}(\vq^\top\rvh)$, a step \emph{up} the variance. To first order in $\eta$ the variance changes by
\begin{equation*}
    \Delta\operatorname{Var}(\vq^\top\rvh)
    =\nabla_\Theta\operatorname{Var}^\top\Delta\Theta+O(\eta^2)
    =\frac{\eta\gamma}{2(\delta^2+\sigma^2)}\bigl\|\nabla_\Theta\operatorname{Var}(\vq^\top\rvh)\bigr\|_2^2+O(\eta^2).
\end{equation*}
Insert $\|\nabla_\Theta\operatorname{Var}\|_2^2=4\delta^2\|\Cov(\rva_\vq,\ru)\|_2^2$ from Step~4 and $\operatorname{Var}(\vq^\top\rvh)=\delta^2$:
\begin{equation*}
    \Delta\operatorname{Var}(\vq^\top\rvh)=\kappa\,\operatorname{Var}(\vq^\top\rvh)+O(\eta^2),
    \qquad
    \kappa=\frac{2\eta\gamma\,\|\Cov(\rva_\vq,\ru)\|_2^2}{\delta^2+\sigma^2}
    \;\xrightarrow[\delta\to0]{}\;\frac{2\eta\gamma\,\|\Cov(\rva_\vq,\ru)\|_2^2}{\sigma^2}.
\end{equation*}
The variance therefore grows by a fixed fraction of itself per step, which is exponential growth at a rate proportional to $\gamma/\sigma^2$. The drift is first order in $\eta$, does not depend on the batch size, and vanishes as $\delta\to0$, consistent with part~(a). Starting from a seed $\delta_0$, reaching a fixed scale takes on the order of $\log(1/\delta_0)/\kappa$ steps, so the size of the seed matters only logarithmically.

\section{EBReg training procedure}
\label{app:ebreg-training}

Algorithms~\ref{alg:ebreg-training} and~\ref{alg:score-tta} detail the EBReg training procedure and score-guided test-time adaptation described in Sections~\ref{sec:ebreg-algorithm} and~\ref{sec:score-tta}, respectively.

\begin{algorithm}[H]
\caption{EBReg training}
\label{alg:ebreg-training}
\small
\textbf{Input:} normalized encoder--projector $g_\vphi\circ f_\vtheta$, noise predictor $\hat{\vepsilon}_\vpsi$, augmentation distribution $\mathcal{T}$; $B,T,K_{\mathrm{score}},\sigma,\gamma$.\par
\textbf{Notation:} $\operatorname{sg}=\operatorname{stopgrad}$. Score-fitting and regularization means pool the $2B$ embeddings; alignment uses the $B$ matched pairs as in \Eqref{eq:ebreg-objective}. Each noise draw is independent and standard Gaussian.\par
\begingroup
\setlength{\tabcolsep}{3pt}
\renewcommand{\arraystretch}{1.1}
\begin{tabular}{@{}r@{\quad}p{0.92\linewidth}@{}}
1 & \textbf{for} $r=1,\ldots,T$ \textbf{do} \\
2 & \quad Sample a minibatch and paired views $\vx^{(1)},\vx^{(2)}$ using $\mathcal{T}$. \\
3 & \quad $\bar\vh\gets\operatorname{sg}(g_\vphi(f_\vtheta(\vx^{(1:2)})))$. \\
4 & \quad \textbf{repeat} $K_{\mathrm{score}}$ \textbf{times} \\
5 & \qquad Draw noise $\vepsilon$; $\tilde\vh\gets\bar\vh+\sigma\vepsilon$. \\
6 & \qquad Update only $\vpsi$ on $\operatorname{mean}\|\hat{\vepsilon}_\vpsi(\tilde\vh)-\vepsilon\|_2^2$. \\
7 & \quad \textbf{end repeat} \\
8 & \quad Recompute $\vh\gets g_\vphi(f_\vtheta(\vx^{(1:2)}))$ on the same views with gradients; draw fresh $\vepsilon$. \\
9 & \quad $\tilde\vh\gets\vh+\sigma\vepsilon$; $\vs\gets\operatorname{sg}(-\hat{\vepsilon}_\vpsi(\tilde\vh)/\sigma)$. \\
10 & \quad Update only $\vtheta,\vphi$ on $\mathcal{L}_{\mathrm{align}}+\gamma\operatorname{mean}(\vs^\top\tilde\vh)$, using \Eqref{eq:ebreg-objective}. \\
11 & \textbf{end for}; \textbf{return} $g_\vphi\circ f_\vtheta$ and $\vs_{\vpsi,\sigma}$. \\
\end{tabular}
\endgroup
\end{algorithm}

\begin{algorithm}[H]
\caption{Score-guided test-time adaptation}
\label{alg:score-tta}
\small
\textbf{Input:} EBReg-trained encoder $f_\vtheta$, fixed projector $g_\vphi$,
score model $\vs_{\vpsi,\sigma}$, classifier $c$;
$K_{\mathrm{tta}}$ and adaptation learning rate $\eta$.\par
\textbf{Notation:} $\valpha\subset\vtheta$ denotes the encoder's normalization
parameters and $\operatorname{sg}=\operatorname{stopgrad}$.
The projector, score model, classifier, and all encoder parameters outside
$\valpha$ remain fixed.\par

\begingroup
\setlength{\tabcolsep}{3pt}
\renewcommand{\arraystretch}{1.1}
\begin{tabular}{@{}r@{\quad}p{0.92\linewidth}@{}}
1 & \textbf{for} each test minibatch $\vx$ \textbf{do} \\
2 & \quad Reset $\valpha$ to its pretrained source values. \\
3 & \quad \textbf{repeat} $K_{\mathrm{tta}}$ \textbf{times} \\
4 & \qquad Compute encoder representation
    $\vr_{\valpha}\gets f_{\vtheta,\valpha}(\vx)$. \\
5 & \qquad Compute score-space embedding
    $\vh_{\valpha}\gets g_\vphi(\vr_{\valpha})$. \\
6 & \qquad $\vs\gets
    \operatorname{sg}\!\left(
    \vs_{\vpsi,\sigma}(\vh_{\valpha})
    \right)$. \\
7 & \qquad $\mathcal{L}_{\mathrm{TTA}}
    \gets -\operatorname{mean}
    (\vs^\top\vh_{\valpha})$. \\
8 & \qquad Update only $\valpha$ on
    $\mathcal{L}_{\mathrm{TTA}}$ with learning rate $\eta$. \\
9 & \quad \textbf{end repeat} \\
10 & \quad Recompute
     $\vr_{\valpha}\gets f_{\vtheta,\valpha}(\vx)$ and predict
     $\hat{\vy}\gets c(\vr_{\valpha})$. \\
11 & \textbf{end for} \\
\end{tabular}
\endgroup
\end{algorithm}

\section{Implementation Details}
In this section, we elaborate on various hyperparameter and design choices for our experiments.
We use ImageNet-1k for all our experiments, using the standard train, validation sets comprising 1.28M and 50K images, respectively. We use ResNet-50, ViT-H, and ViT-L as the major architectural choices for our study. The augmentation pipeline matches those used in the corresponding protocols. We use a simple MLP as a denoiser in all the cases. We use a simple MLP projector for both VICREG and LeJEPA cases with projector dims of 256 and 64, respectively. Other hyperparameters for each of the protocols are listed below:

\paragraph{VICReg protocol.}
The projector has layer widths $8192$--$8192$--$256$. The denoiser uses a hidden width of 2048 and four blocks. The alignment-loss coefficient is 1.

\paragraph{LeJEPA protocol.}
The augmentation pipeline produces two global crops of size $224\times224$ with scale range $[0.3,1.0]$ and eight local crops of size $98\times98$ with scale range $[0.05,0.3]$. Images use ImageNet channel-wise normalization with mean $(0.485,0.456,0.406)$ and standard deviation $(0.229,0.224,0.225)$. The learning-rate warmup starts at $0.01$ times the configured learning rate. Training uses bfloat16 precision.

\paragraph{I-JEPA/StoP protocol.}
The input crop size is $224\times224$, with crop scale range $[0.3,1.0]$ and patch size 14. We use one context mask with scale range $[0.85,1.0]$ and four target masks with scale range $[0.15,0.2]$, mask aspect ratios in $[0.75,1.5]$, and a minimum of 10 retained patches. The predictor has depth 12 and embedding dimension 384. The initial and final learning rates are $2\times10^{-4}$ and $10^{-6}$, respectively; the configuration specifies constant learning-rate and weight-decay strategies from epoch 250. StoP is enabled with variance parameter $0.2$.

\section{Maximum-entropy targets under each constraint}
\label{app:maxent-connections}

We compute the maximum-entropy distribution under each constraint of Table~\ref{tab:maxent-targets}, labelled (i)--(iv) in the order of the table. All four follow from one standard computation.

\paragraph{The maximum-entropy recipe.}
Consider distributions on a support $\mathcal{S}$ that satisfy moment constraints $\E[f_k(\rvh)]=c_k$ for $k=1,\ldots,m$. Suppose the exponential-family density
\begin{equation}
    p^\star(\vh)=\frac1Z\exp\Bigl(-\sum_{k=1}^m\lambda_kf_k(\vh)\Bigr)\quad\text{on }\mathcal{S}
    \label{eq:maxent-recipe}
\end{equation}
satisfies the constraints for some multipliers $\lambda_k$. Then $p^\star$ is the unique maximum-entropy distribution. To see this, take any feasible $p$. Because $-\log p^\star=\log Z+\sum_k\lambda_kf_k$, its average under $p$ involves only the constrained moments:
\begin{equation*}
    -\int p\log p^\star=\log Z+\sum_k\lambda_k\,\E_p[f_k(\rvh)]=\log Z+\sum_k\lambda_kc_k=H(p^\star).
\end{equation*}
Hence
\begin{equation*}
    H(p)=-\int p\log p^\star-D_{\mathrm{KL}}(p\,\|\,p^\star)=H(p^\star)-D_{\mathrm{KL}}(p\,\|\,p^\star)\leq H(p^\star),
\end{equation*}
with equality only when $p=p^\star$. With no moment constraints, $m=0$, the recipe gives the constant density, that is, the uniform distribution on $\mathcal{S}$.

\paragraph{(i) Quadratic constraint: isotropic Gaussian.}
Let $\E[\rvh]=\vzero$ and $\operatorname{tr}\Cov(\rvh)=d$; unit marginal variances are a special case. The noisy embedding $\rvz=\rvh+\sigma\rvepsilon$ then satisfies
\begin{equation*}
    \E[\rvz]=\vzero,
    \qquad
    \E\|\rvz\|_2^2=\E\|\rvh\|_2^2+\sigma^2\E\|\rvepsilon\|_2^2=d(1+\sigma^2).
\end{equation*}
Apply the recipe to $\rvz$ on $\mathcal{S}=\R^d$ with $f(\vz)=\|\vz\|_2^2$. It gives $p^\star(\vz)\propto\exp(-\lambda\|\vz\|_2^2)$, a zero-mean isotropic Gaussian, and matching the second moment fixes its variance at $1+\sigma^2$ per coordinate. Its entropy is the bound
\begin{equation}
    H_\sigma(\rvh)=H(\rvz)\leq H\bigl(\mathcal{N}(\vzero,(1+\sigma^2)\mI_d)\bigr)=\frac d2\log\bigl(2\pi e(1+\sigma^2)\bigr),
    \label{eq:gaussian-maxent}
\end{equation}
with equality only when $\rvz\sim\mathcal{N}(\vzero,(1+\sigma^2)\mI_d)$. This happens exactly when $\rvh\sim\mathcal{N}(\vzero,\mI_d)$: adding the independent noise $\sigma\rvepsilon$ to $\mathcal{N}(\vzero,\mI_d)$ gives that Gaussian, and conversely the characteristic function of $\rvz$ is the product $\varphi_\rvh({\bm{\omega}})\exp(-\frac12\sigma^2\|{\bm{\omega}}\|_2^2)$, so dividing by the noise factor leaves $\varphi_\rvh({\bm{\omega}})=\exp(-\frac12\|{\bm{\omega}}\|_2^2)$. The result holds for every $\sigma>0$. Since the maximizer has $\Cov(\rvh)=\mI_d$, a rank-deficient representation is never entropy-maximal under this constraint. The target $\mathcal{N}(\vzero,\mI_d)$ is SIGReg's reference~\citep{balestriero2025lejepaprovablescalableselfsupervised}, its covariance is the second-moment condition that VICReg's penalties drive toward~\citep{bardes2022vicreg}, and batch normalization without affine parameters enforces the constraint up to finite-batch error.

\paragraph{(ii) Spherical constraint: uniform on the sphere.}
Let $\|\rvh\|_2=1$, so $\mathcal{S}=\mathbb{S}^{d-1}$ and there is no moment constraint. The recipe gives the constant density with respect to surface measure, the uniform distribution, with entropy $\log|\mathbb{S}^{d-1}|$. The same holds on a sphere of any radius, such as $\|\rvh\|_2=\sqrt d$ under RMS normalization. Layer normalization additionally centers each embedding, $\vone^\top\rvh=0$, so $\mathcal{S}$ is the sphere inside the hyperplane orthogonal to $\vone$ and the target is uniform on that lower-dimensional sphere. Spherical uniformity is the target that contrastive objectives approach through negatives~\citep{wang2020understanding}.

\paragraph{(iii) Categorical constraint: uniform code usage.}
Let $\rvh\in\{\ve_1,\ldots,\ve_K\}$, so $\mathcal{S}$ is a set of $K$ points with counting measure and again no moment constraint. The recipe gives equal probabilities $1/K$, with entropy $\log K$. Uniform usage is SwAV's equipartition constraint and the marginal-entropy regularizer of MSN~\citep{caron2020swav,assran2022msn}; DINO's centering and sharpening act as a related mechanism rather than a prescribed density~\citep{caron2021dino}. Marginal balance across inputs differs from uninformative uniform predictions for every input.

\paragraph{(iv) Moment constraint: generalized Gaussian.}
Let $\E\|\rvh\|_p^p=M$ on a support $S$. The recipe with $f(\vh)=\|\vh\|_p^p$ gives
\begin{equation}
    p^\star(\vh)=Z^{-1}\exp\bigl(-\alpha\|\vh\|_p^p\bigr)\mathbb{1}_{S}(\vh),
    \label{eq:generalized-gaussian-target}
\end{equation}
with $\alpha$ chosen to satisfy the constraint, whenever this density is normalizable. On $\R^d$ it is a product generalized Gaussian, which reduces to the Gaussian of part~(i) at $p=2$, and restricting to the positive orthant gives its truncation. This is the principle behind Rectified LpJEPA's target design~\citep{kuang2026rectifiedlpjepajointembeddingpredictive}, whose additional rectification introduces atoms at zero and requires mixed-measure entropy.

\paragraph{Effect of the smoothing.}
Part~(i) is computed directly for the smoothed entropy $H_\sigma$ and is exact at every noise level. Parts~(ii)--(iv) are computed for the entropy of $\rvh$ itself, which $H_\sigma$ approaches from above as $\sigma\to0$: adding independent noise can only add entropy, $H_\sigma(\rvh)\geq H(\rvh)$, and the excess vanishes with $\sigma$. For the sphere and the codes the targets remain maximizers of $H_\sigma$ at every $\sigma>0$ by symmetry. Rotating a distribution on the sphere, or permuting the codes, leaves $H_\sigma$ unchanged because the noise is isotropic; the uniform distribution is the average of all such rotated or permuted copies; and the entropy of a mixture is at least the average entropy of its components. For the moment constraint with $p\neq2$ the maximizer of $H_\sigma$ at fixed $\sigma>0$ is a smoothed analogue of \Eqref{eq:generalized-gaussian-target} that we do not characterize.

\section{Use of Large Language Models}
\label{app:llm-usage}
LLMs were used to assist with writing, revising, organizing, and polishing the manuscript; coding tasks; table and bibliography formatting; and LaTeX troubleshooting. They also assisted in deriving proofs presented in the appendix, reviewing mathematical arguments, identifying gaps between stated assumptions and claimed conclusions, and drafting revisions to narrow the interpretation of the directional gradient results. All experimental values used in preparing tables and discussing results were supplied by the authors. The authors take full responsibility for the manuscript’s correctness, originality, and proper attribution, including all LLM-assisted text, code, proofs, and mathematical reasoning.

\end{document}